\documentclass[letterpaper, 10 pt, conference]{ieeeconf}
\usepackage{amssymb}  
\usepackage{algorithm}
\usepackage{algpseudocode}
\usepackage{amsmath}
\usepackage{array}  
\usepackage{cite}
\usepackage{fancyhdr}

\usepackage{etoolbox}
\usepackage{accents}
\usepackage{xcolor}
\usepackage{hyperref}
\usepackage{placeins}
\usepackage{multirow}
\usepackage{booktabs}
\usepackage{amsmath, amssymb}
\usepackage{mathrsfs}
\usepackage{pdfpages}
\usepackage{wrapfig}
\newcommand{\specialcell}[2][c]{%
  \begin{tabular}[#1]{@{}c@{}}#2\end{tabular}}

\title{\LARGE \bf  Impedance Primitive-augmented Hierarchical Reinforcement Learning for Sequential  Tasks}
\author{ Amin~Berjaoui~Tahmaz$^\dagger$, Ravi Prakash$^\ddagger$, Jens Kober$^\dagger$ 
\thanks{ 
Authors $^\dagger$ are with TU Delft, Netherlands.
Author $^\ddagger$ is with IISc Bangalore, India.
{\tt\small amine.berjawi123@gmail.com, ravipr@iisc.ac.in, j.kober@tudelft.nl}%
}}

\IEEEoverridecommandlockouts 
\usepackage[]{eso-pic}

\newcommand{\placetextbox}[3]{
  \setbox0=\hbox{#3}
  \AddToShipoutPictureFG*{
    \put(\LenToUnit{#1\paperwidth},\LenToUnit{#2\paperheight}){\vtop{{\null}\makebox[0pt][c]{#3}}}%
  }%
}%

\begin{document}
\maketitle
\placetextbox{0.5}{0.97}{\textcolor{blue}{This article is accepted for publication in IEEE International Conference on Robotics and Automation (ICRA) 2025. Copyright @ IEEE.}}%

\begin{abstract}
    This paper presents an Impedance Primitive-augmented hierarchical reinforcement learning framework for efficient robotic manipulation in sequential contact tasks. We leverage this hierarchical structure to sequentially execute behavior primitives with variable stiffness control capabilities for contact tasks. Our proposed approach relies on three key components: an action space enabling variable stiffness control, an adaptive stiffness controller for dynamic stiffness adjustments during primitive execution, and affordance coupling for efficient exploration while encouraging compliance. Through comprehensive training and evaluation, our framework learns efficient stiffness control capabilities and demonstrates improvements in learning efficiency, compositionality in primitive selection, and success rates compared to the state-of-the-art. The training environments include block lifting, door opening, object pushing, and surface cleaning. Real world evaluations further confirm the framework's sim2real capability. This work lays the foundation for more adaptive and versatile robotic manipulation systems, with potential applications in more complex contact-based tasks.
\end{abstract}



\section{Introduction}

    Realistic manipulation tasks involve a prolonged sequence of motor skills in varying environments. For decades, the challenge of enabling robotic manipulators to solve realistic long-horizon tasks has persisted. While existing research has made strides in addressing important aspects of long-horizon tasks, a critical gap remains in the context of contact-rich environments, highlighting a crucial area that requires further exploration and development. An example can be found in a common manipulation task: object sorting. A robot should be able to plan a series of precise actions over time while adjusting its positioning and applied forces to accommodate objects of varying shapes and sizes while also taking the interaction environment into consideration. This paper focuses on the intersection of deep reinforcement learning (DRL) and adaptive stiffness control to address this longstanding challenge.

Prior works have extensively explored robotic manipulation in long-horizon applications. Conventional methods often use state machines \cite{ranjbar2021residual}\cite{li2012object} or symbolic reasoning \cite{nguyen2021self}\cite{zhao2021sydebo} to learn action sequences for solving a task. However, these approaches explicitly design the decision-making sequence, which may introduce constraints that limit adaptability to different tasks and contribute to error accumulation throughout the task sequence. In response to these limitations, learning techniques such as hierarchical reinforcement learning (HRL) \cite{botvinick2012hierarchical} have been employed, establishing themselves as a common approach for problems requiring sequential decision-making.

When deploying long-horizon frameworks in contact-rich environments, the integration of stiffness control becomes crucial for adapting to external forces and uncertainties during task execution. This adaptability ensures precision and stability in navigating contact-rich environments. However, despite a substantial body of research dedicated to variable stiffness control, current approaches are primarily tailored to short-horizon applications. These methods typically involve designing controllers that adjust end-point force in response to environmental forces \cite{franklin2007endpoint}, adapting impedance and damping parameters through learning techniques \cite{johannsmeier2019framework}\cite{martin2019variable}, and learning from a human demonstrator \cite{abu2018force}\cite{dou2022robot}.

\begin{figure}
  \includegraphics[width=\columnwidth]{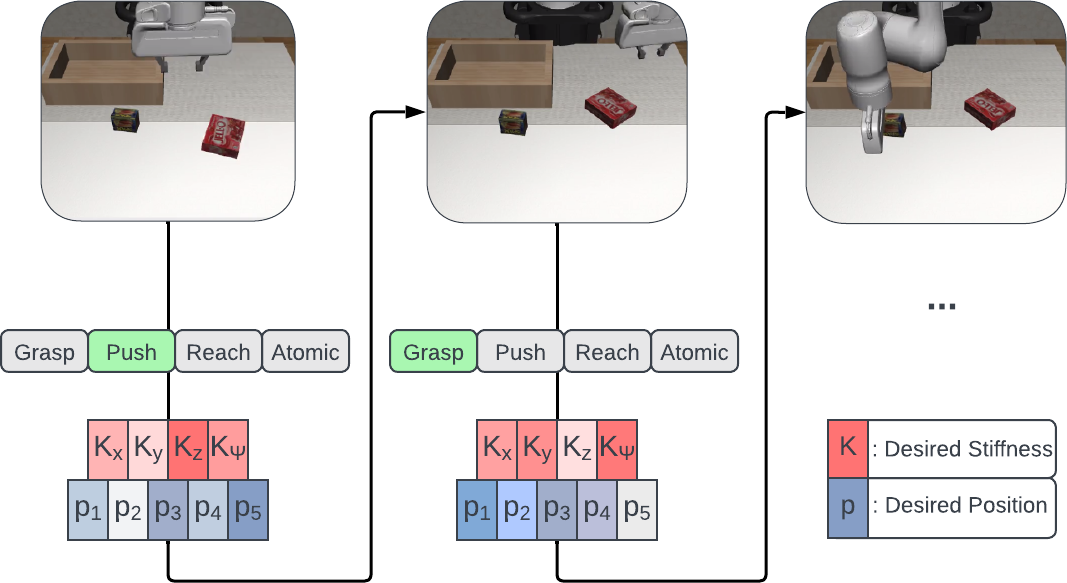}
  \caption{Figure shows the augmentation of the impedance primitive into HRL policy. 
  }
  \label{fig:imp-hrl}
\end{figure}

This paper aims to bridge the gap between sequential task planning and adaptive stiffness control using a DRL framework. We design an HRL framework, as shown in Figure \ref{fig:imp-hrl}, that selects a high-level action primitive from a pre-defined library and outputs an initial estimate for controller parameters for low-level control. During primitive execution, an adaptive controller is initiated to optimize the robot's stiffness, aiming for an balance between safety (reducing interaction forces with the environment) and performance (ensuring task completion). This design allows the robot to dynamically optimize stiffness parameters, enabling it to transition between high stiffness for precision tasks and increased compliance for enhanced adaptability. We present experiments conducted in both simulation and the real world, focusing on sequential tasks that deal with different contact challenges. Our results highlight notable advantages when compared to a state-of-the-art baseline.

The remainder of this paper is structured as follows: Section~\ref{sec:related_works} discusses related work. Section~\ref{sec:problem_statement} defines the problem statement. Section~\ref{sec:preliminaries} provides necessary background and preliminaries. Our proposed  Impedance Primitive-augmented HRL (IMP-HRL) approach is introduced in Section~\ref{sec:imp_hrl}. Section~\ref{sec:experiments} presents experimental results and analysis. 
Section~\ref{sec:conclusion} concludes the paper with a summary of findings and future research directions.
\section{Related Works}
\label{sec:related_works}
\textit{Sequential Planning:} Extensive work in task and motion planning (TAMP) spans various robotics applications, involving explicit decision-making frameworks and machine learning for learned behavior sequences.
Common approaches employ hierarchical task planning, combining high-level planners with low-level controllers. In robot manipulation, this often takes the form of finite state machines \cite{ranjbar2021residual, gal2021state, onishi2023priority} or behavior trees \cite{french2019learning, rovida2017extended} as high-level controllers. Similar methods use symbolic reasoning \cite{cheng2022guided, agia2022taps, wu2021example}, representing high-level tasks and constraints with symbols. Although these methods offer explainability, their pre-defined nature limits adaptability to real-world variability, leading to suboptimal performance. Our proposed framework addresses this by learning the high-level planner and optimizing low-level controller parameters for better generalization and robustness.
Recently, learning approaches have emerged to overcome these limitations. Imitation learning (IL) is a key candidate for sequential planning, enabling robots to learn demonstrated behavior sequences. Behavior cloning, a well-established IL method, has robots replicate demonstration sequences \cite{liu2020understanding, zhang2018deep, wu2020squirl}, but this limits generalizability. Advanced IL methods aim to generalize learned sequences \cite{liang2022learning, mandlekar2020learning, huang2019neural}, yet they still struggle to adapt to new environments. Our framework adapts action sequences to the environment state, addressing this limitation and mitigating suboptimal performance from human error in demonstration data.
Hierarchical Reinforcement Learning (HRL) has gained attention for long-horizon planning. State-of-the-art approaches like MAPLE \cite{nasiriany2022augmenting}, RAPS \cite{dalal2021accelerating}, and STAP \cite{agia2023stap} train hierarchical policies to choose and execute primitives from a behavior library. Despite handling complex tasks and improving sample efficiency, these methods rely on static controllers, which hinder performance in contact tasks and pose risks in real-world settings. Our method builds on these concepts, optimizing stiffness to maximize compliance without compromising task success.

\textit{Variable Stiffness Control:} Existing methods for adapting the stiffness of an impedance controller typically use task-specific impedance profiles. Common approaches include learning from demonstration methods, such as Dynamic Motion Primitives \cite{zhou2016learning, nemec2013transfer, pastor2009learning} or Gaussian Mixture Models \cite{abu2018force, cederborg2010incremental}. Alternatively, some methods schedule variable stiffness gains for different task phases \cite{li2018force, mitrovic2011learning, ruckert2013learned}. Despite their ease of application, these methods struggle to generalize stiffness profiles across tasks and depend on expert demonstrators.
RL has emerged as a promising method for learning stiffness profiles. Some methods bootstrap the RL policy with initial stiffness demonstrations \cite{theodorou2010generalized, rey2018learning, kim2022scape} to accelerate learning, which are then optimized for specific tasks. However, the reliance on expert demonstrators remains an issue. Other RL approaches focus on designing an appropriate action space in which an agent samples impedance parameters as actions to adapt controller behavior. For adaptive stiffness applications, an impedance action space allows the agent to learn stiffness and damping parameters in joint space \cite{bogdanovic2020learning} and end-effector space \cite{martin2019variable}. Similar approaches use residual reinforcement learning, where a policy outputs actions to support an existing controller \cite{ranjbar2021residual, beltran2020learning, kulkarni2022learning}. However, these methods fail in long-horizon tasks due to their limited ability to capture sequential dependencies.

\textbf{Contributions:} The main contributions of this work are i)  an  impedance primitive augmented HRL framework for sequential contact tasks, ii)  a novel behavior affordance that concurrently optimizes for position and compliance; (iii) an adaptive controller for dynamic stiffness modifications for optimal  execution in varying environments.

\section{Problem Statement}
\label{sec:problem_statement}
    The long-horizon robotics manipulation task can be formulated within the framework of HRL combined with Parameterized Action Markov Decision Processes (PAMDPs) \cite{masson2016reinforcement}.
Let \( S \) represent the state space, and \( \pi_H: S \rightarrow A_H \) be the high-level policy that selects high-level actions \( a_H \in A_H \), which define primitives. For each high-level action \( a_H \), let \( \pi_{param} :S\times \mathcal{L}  \rightarrow  \Theta \) be the corresponding low-level policy that selects parameterized actions \( (p, \theta) \). The overall policy \( \pi(s) = \pi_{param}^{\pi_H(s)}(s) \) determines the hierarchical decision-making process. The environment dynamics are captured by the transition function \( P(s' | s, \bar{a}) \) and reward function \( R(s, \bar{a}) \), where \( \bar{a} = (p, \theta) \). The objective is to find the hierarchical policy \( \pi \) that maximizes the expected cumulative reward \( J(\pi) = \mathbb{E} \left[ \sum_{t=0}^{\infty} \gamma^t r_t \right] \), optimizing both the high-level task decomposition and the execution of parameterized actions for efficient manipulation.


\section{Preliminaries: MAPLE \cite{nasiriany2022augmenting}}
\label{sec:preliminaries}

MAPLE \cite{nasiriany2022augmenting} is a state-of-the-art HRL framework that frames the sequential decision making problem as a PAMDP. It uses a two-level policy structure: a high-level task policy $\pi_H$ selects a behavior primitive from a library $L = \{p_1, p_2, ..., p_n\}$, while a low-level parameter policy $\pi_L^{aH}$ predicts the parameters $\theta$ for the chosen primitive. Each primitive executes a closed-loop control sequence, minimizing the error between the current state $s$ and the target state $\theta$. The  primitives and their parameters are documented in Table~\ref{tab:primitives}.

To enhance exploration, MAPLE incorporates position affordances, which are rewards that encourage interactions near task-relevant objects. This position affordance is modeled as 
\begin{equation}\label{eq:pos_afford}
a_{\text{pos}}(s, \theta; p) = \max_{\kappa \in \mathcal{K}} \bigl(1 - \tanh\bigl(\max\left(||\theta- \kappa|| - \tau, 0\right)\bigr)\bigr),
\end{equation}
where $\mathcal{K}$ represents the set of object keypoints and $\theta$ is the chosen parameters for a primitive. Accordingly, the affordance reward increases as it approaches objects in the environment.

MAPLE’s structured approach facilitates learning of parameterized skills but lacks explicit mechanisms for impedance control, which is critical for contact-rich tasks. Our proposed \emph{IMP-HRL} extends MAPLE by integrating impedance primitives and adaptive stiffness control, enabling more robust interaction with the environment.


\begin{table}[t!]
    \centering
    \caption{Description of primitives and their parameters}
    \begin{tabular}{>{\centering\arraybackslash}m{1.8cm}>{\centering\arraybackslash}m{3.5cm}>{\centering\arraybackslash}m{2cm}}
        \toprule
        \textbf{Primitive} & \textbf{Description} & \textbf{Parameters} \\
        \midrule
        Reach & Moves the end-effector to a target location & \shortstack[t]{$(x, y, z)$} \\
        \midrule
        Grasp & Moves end-effector to grasp location then activates gripper & \shortstack[t]{$(x, y, z, \psi)$} \\
        \midrule
        Push & Moves end-effector to a target location, then applies a displacement $\delta$ & \shortstack[t]{$(x, y, z,$\\ $\delta_x, \delta_y, \delta_z)$} \\
        \addlinespace 
        \midrule
        Atomic & Apply atomic action & \shortstack[t]{$(\delta_x, \delta_y, \delta_z)$} \\
        \midrule
        Gripper & Open/Close binary gripper & $g$ \\
        \bottomrule
    \end{tabular}
    \label{tab:primitives}
\end{table}

\section{IMP-HRL}
\label{sec:imp_hrl}

We propose Impedance Primitive-augmented HRL (IMP-HRL) for robust sequential contact tasks. We introduce two components into the MAPLE framework that allow us to to achieve variable impedance control for sequential contact tasks.

    
    

\subsection{Impedance Primitive}\label{subsec:action-space}

To accommodate contact-rich environments, the target states need to be extended from exclusively position-based parameters as in MAPLE to also include variable impedance parameters.
We propose augmenting HRL with the  primitive parameter action space  containing the  position and impedance  parameters \cite{martin2019variable}. It allows the agent to control the impedance parameters by sampling them as actions. This augmentation extends the parameter  space, $\theta$, to now contain $(K_x,K_y,K_z)$ for variable stiffness/impedance control along different coordinate axes and $K_{\psi}$ for handling orientation or angular variations (shown in Figure \ref{fig:imp-hrl}). The damping term $D$ in the impedance parameters are selected based on critical damping  of system's closed loop response to reduce the number of learnable parameters. 

A limitation of this primitive representation arises from the sequential nature of decision-making: once the policy triggers a behavior primitive, it is required to wait for the primitive to complete its execution before modifying the stiffness value again. On the other hand, using an action space with dynamically adapting stiffness parameters introduces a learning challenge. Therefore, the stiffness parameters predicted by the parameter policy will act as an initial stiffness prediction which will be further adjusted using an adaptive stiffness controller.

\begin{figure}[b!]
  \centering
   \includegraphics[width=.7\columnwidth]{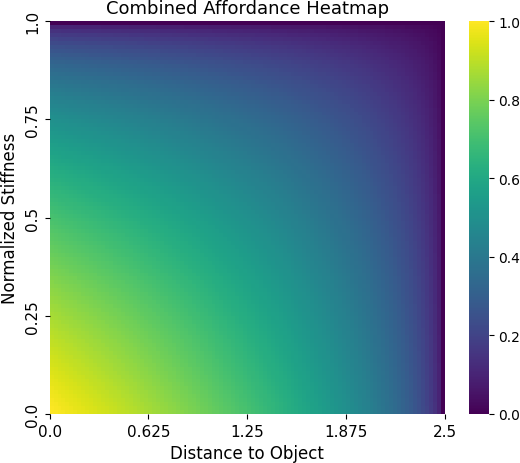}
  \caption{Heatmap visualization of affordance coupling}
  \label{fig:heatmap}
\end{figure}

\textbf{Affordance Coupling - Combining Position and Stiffness Affordances:} In the context of tasks that can benefit from stiffness control, position-based affordances \eqref{eq:pos_afford} are insufficient since they focus exclusively on spatial information. To address this limitation, we propose an additional \emph{stiffness affordances} to maximize compliance whenever possible. In turn, this translates to a reduction in interaction forces between the robot and the environment, which improves the overall safety of the system. Accordingly, stiffness is only increased when it is necessary to meet task requirements. This stiffness affordance is modeled as

This coupling model improves exploration efficiency and encourages the agent to select low-stiffness parameters during the early stages of training. Furthermore, this method eliminates the necessity for careful reward weight tuning that is typically required when directly penalizing high stiffness values. Such tuning would otherwise need to be conducted for each new environment, potentially having a detrimental effect on learning performance \cite{faust2019evolving}.
Note that the \emph{atomic} and \emph{gripper release} always have an affordance of 1 due to their general utility.


\begin{figure}[b]
  \centering
  \includegraphics[width=.9\columnwidth]{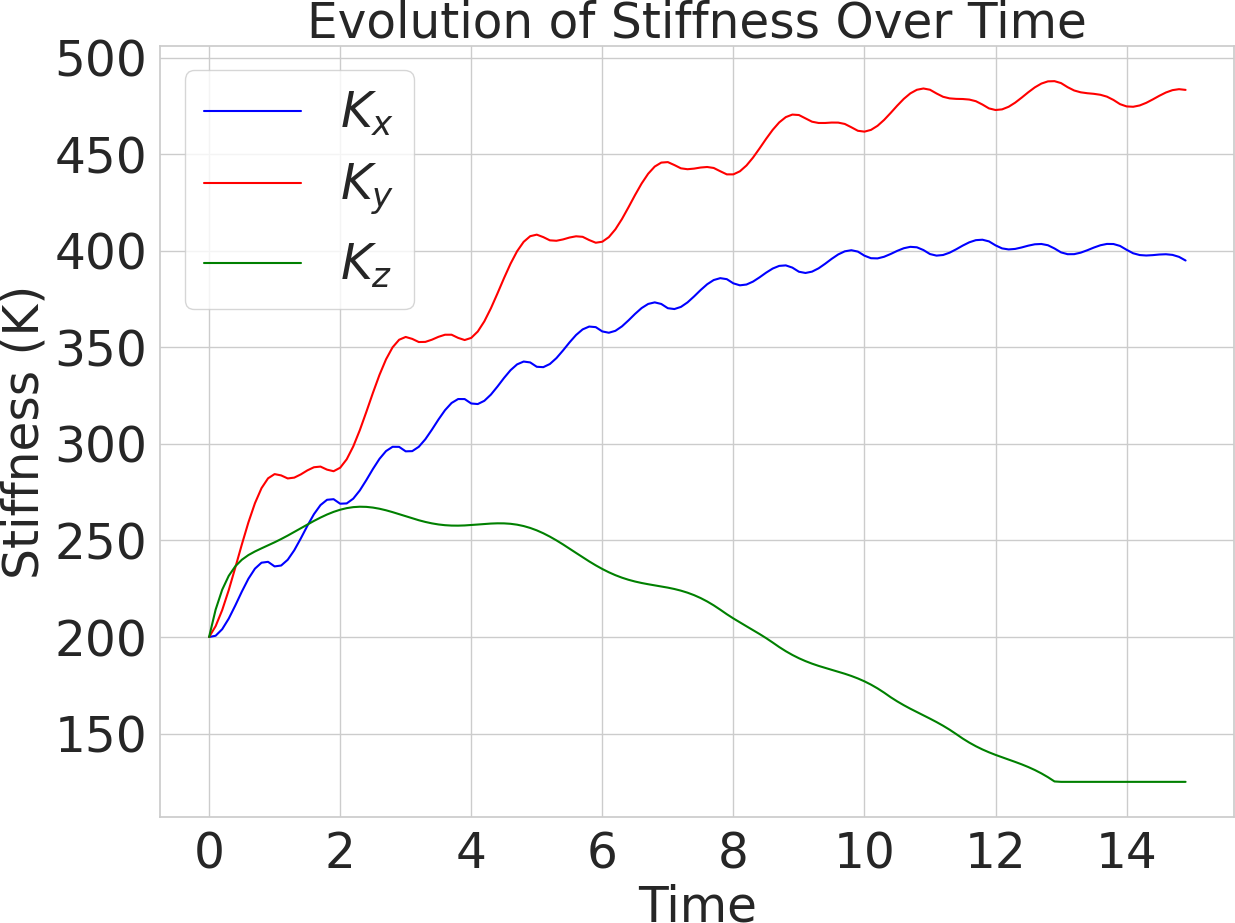}
    \caption{Example of adaptive stiffness when wiping.}
    \label{fig:aforce-stiffness-sample}
\end{figure}
\subsection{Adaptive Controller}\label{sec:adaptivecontrol}

\begin{figure}[t!]
  \centering
    \includegraphics[width=.9\columnwidth]{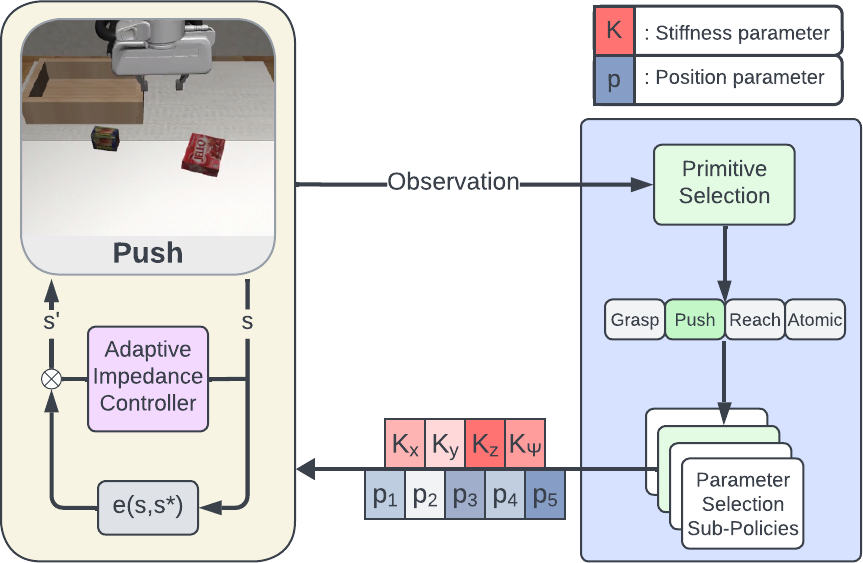}
  \caption{Adaptive impedance controller integrated within the low-level parametrized policy.}
  \label{fig:adap_control}
\end{figure}

After the policy selects a primitive and its parameters, the behavior is executed through a closed loop control scheme. Using the stiffness parameters outputted by the parameter policy as an initial estimate of the required stiffness to complete a given stage of the task, this stiffness is adapted in real-time using an adaptive stiffness controller. Figure \ref{fig:adap_control} shows the adaptive impedance controller integrated within the low level parametrized policy. 

The \textbf{Adaptive Controller} used in this  mimics human muscle stiffness during motion execution \cite{ulmer2021learning} by adapting the stiffness in accordance with the output of
\begin{equation}\label{eq:Kdot}
\dot{K}(t) = \beta|\epsilon(t)| - \gamma E,
\end{equation}
where $\epsilon(t)$ is the closed loop feedback error and $E$ is the energy consumed by the robot joints, while $\beta$ and $\gamma$ scale these values to influence the stiffness behavior. As for the corresponding damping matrix, it satisfies a critical damping condition such that $D(t) = 2\sqrt{K(t)}$, which is re-calculated every time the stiffness value is updated. It is important to note that interpolation is used to generate intermediate points along the trajectory toward a target state, which prevents drastic changes in stiffness.

In practice, the controller stiffness is initialized using the stiffness output of the low-level RL policy. Then, it calculates the stiffness at the next step by using $\beta$ to scale the increase in stiffness proportional to the feedback error $e(s-\theta)$. Simultaneously, it reduces stiffness by scaling current energy consumption $E$ with $\gamma$. This process yields a net increase or decrease in the controller's stiffness. Figure \ref{fig:aforce-stiffness-sample} demonstrates an example in which a robot performs an elliptical wiping motion.

Since primitives are simple linear movements, the values of $\beta$ and $\gamma$ can be obtained by performing kinesthetic demonstrations of the primitives, extracting their stiffness profiles, and minimizing the MSE between the demonstrations and the controller output. This yields $\beta$ and $\gamma$ parameters that closely resemble human stiffness behavior. Alternatively, the controller parameters can be determined by simply tuning them until controller performance is satisfactory. 

\begin{figure}[b!]
    \centering
    \includegraphics[width=0.99\columnwidth]{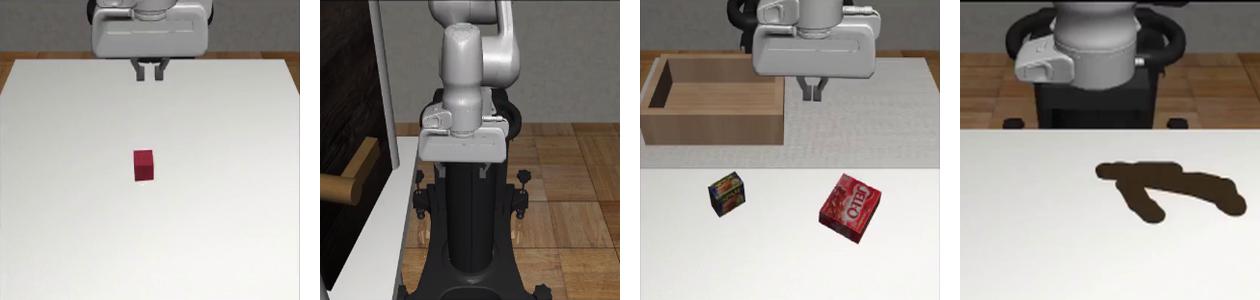}
    \caption{Simulation Experiments: Lift, Door, Cleanup, Wipe}
    \label{fig:rob-sim}
\end{figure}

\begin{figure}[b!]
    \centering
    \includegraphics[width=0.55\columnwidth]{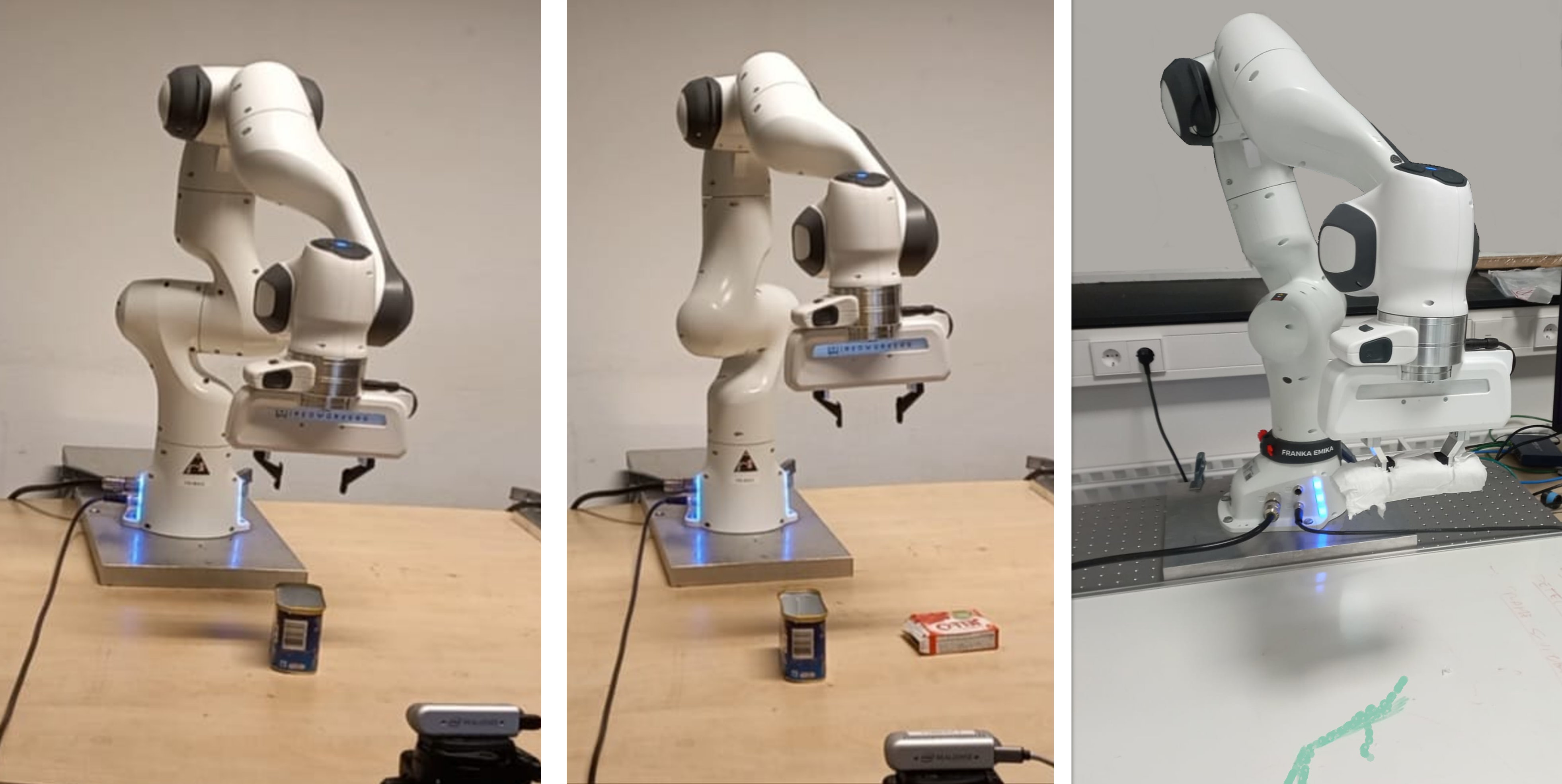}
    \caption{Real Experiments: Lift, Cleanup, Wipe}
    \label{fig:rob-real}
\end{figure}

\begin{figure*}[t!]
    \centering
    \includegraphics[width=.23\textwidth]{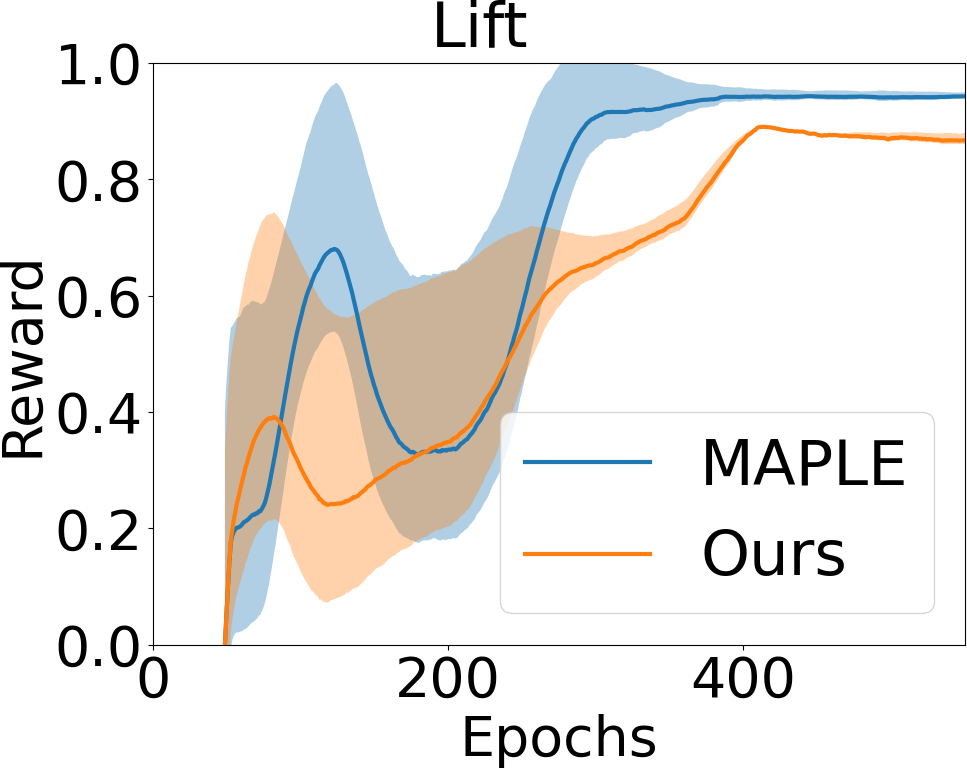}
    \hfill 
    \includegraphics[width=.23\textwidth]{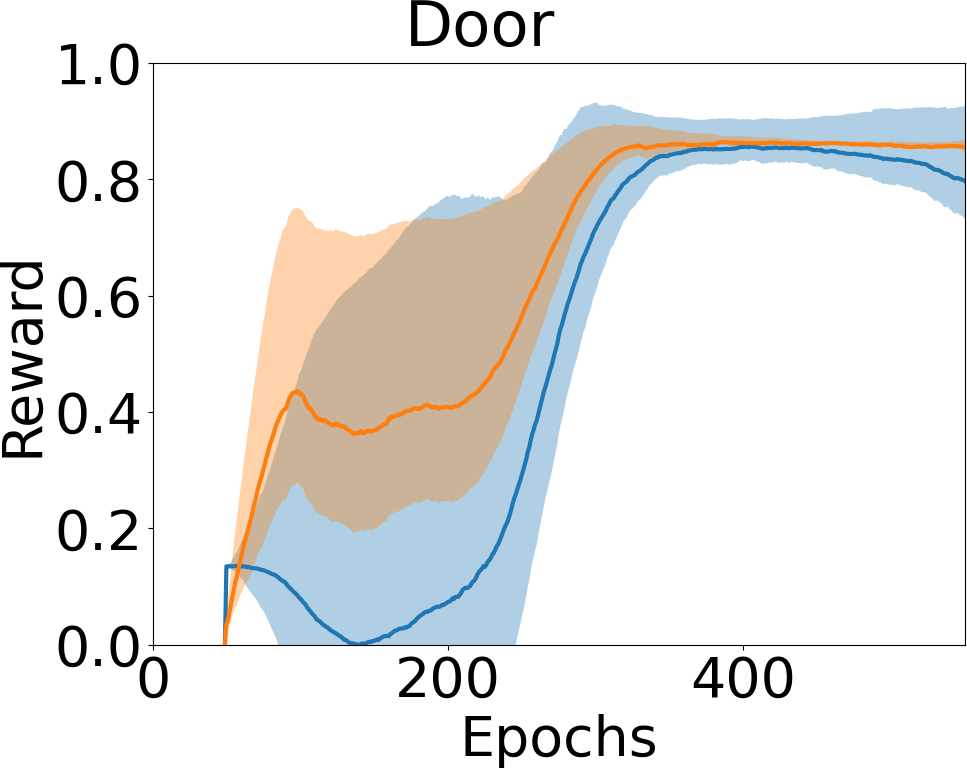}
    \hfill 
    \includegraphics[width=.23\textwidth]{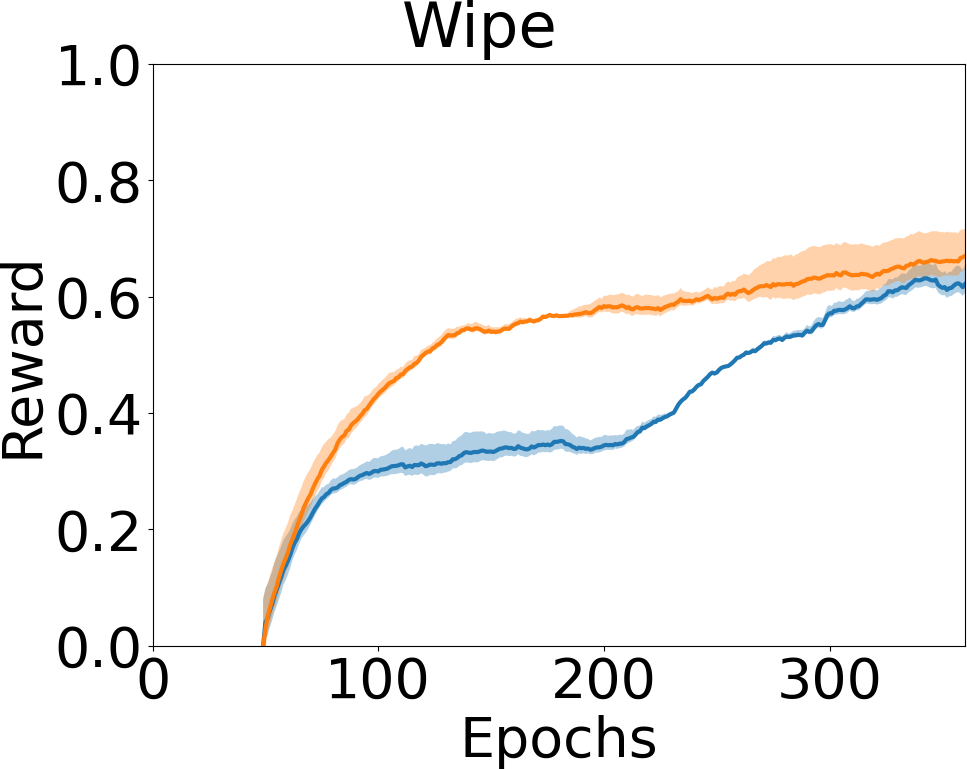}
    \hfill 
    \includegraphics[width=.23\textwidth]{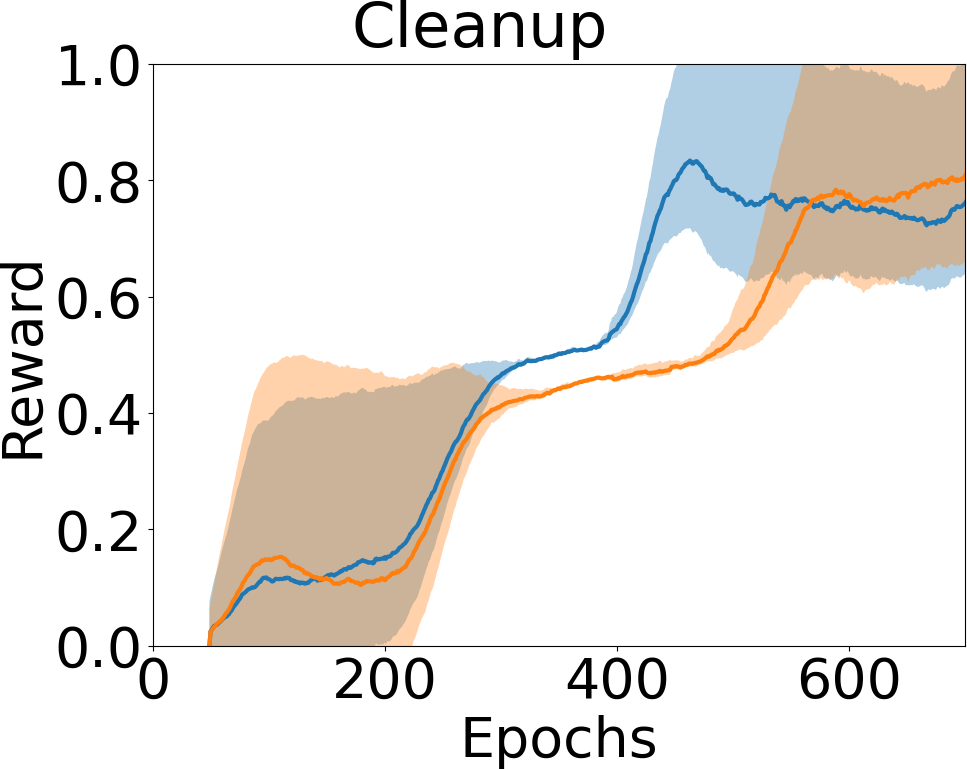}
    \caption{Comparison of learning behavior and convergence times for various tasks. The rewards are averaged over 20 episodes then normalized between 0 and 1 (which represents the maximum reward at each timestep).}
    \label{fig:all-rewards}
\end{figure*}
\begin{figure*}[h!]
    \centering
    \includegraphics[width=\textwidth]{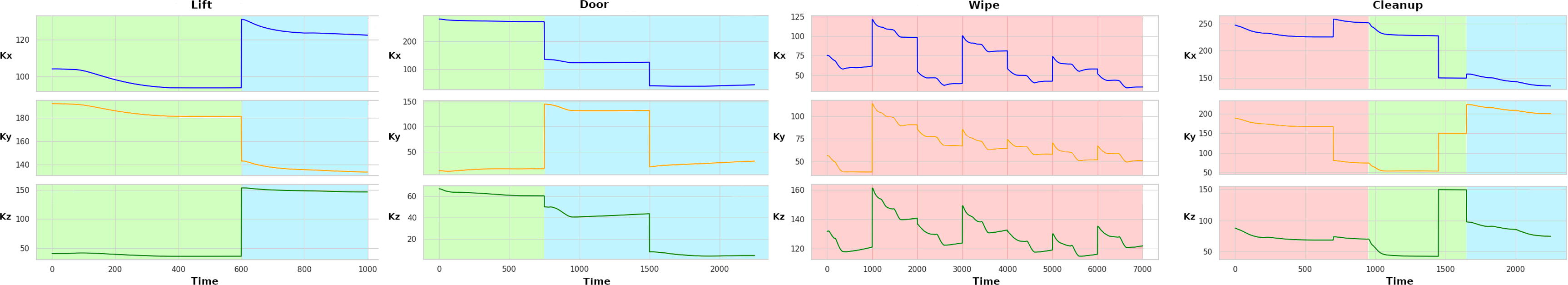}
    \caption{Variable stiffness behavior demonstrating an emphasis on compliance and stiffness reduction. Each background grid colour represents a different primitive being executed - \textcolor{green}{grasp}, \textcolor{cyan}{reach}, \textcolor{red}{push}.}
        \label{fig:stiff_profile}
\end{figure*}
\section{Experimental Results}
\label{sec:experiments}
    In the experiments, we investigated the framework's learning efficiency, analyzed its stiffness and force behavior, highlighted patterns in primitive selection, and evaluated its performance in a real-world setting. This section is divided into  experimental setup,   evaluation in simulation and real robot, and comparative analysis with respect to state-of-the-art method on sequential task execution.


\subsection{Experimental Setup}\label{sec:experimental-setup}

We evaluated our framework in four contact-rich environments: Lift, Door, Wipe, and Cleanup. These interactions include basic object manipulation in the Lift environment, continuous contact in the Door and Wipe environments, and a mix of contact and manipulation interactions in the Cleanup environment. The robot utilized for these experiments was a Franka Emika Panda in the Robosuite simulator \cite{zhu2020robosuite} (see Figure \ref{fig:rob-sim}) and real-world (see Figure \ref{fig:rob-real}).
We additionally apply domain randomization by randomly varying table friction, table height, object positions, and initial end-effector position. Lastly, all the reported results were averaged across 5 random seeds.

\subsection{Comparative Analysis - Simulation}\label{sec:experimental-eval-sim}

We compare our proposed framework with the MAPLE baseline. 
The chosen evaluation metrics are Learning Performance, Maximum Interaction Force, Compositionality, and Success Rate.

\textbf{Evaluation Metrics:} In \emph{Learning Performance}, we examine learning convergence time to assess learning efficiency of the proposed framework. In \emph{Maximum Interaction Force}, we evaluate our framework's ability to adapt stiffness across diferent contexts and its effect on the applied forces. In \emph{Compositionality}, we quantify recurring patterns in primitive selection using a \emph{compositionality metric} \cite{nasiriany2022augmenting}. Lastly, in \emph{Success Rate}, we analyze the framework's ability to consistently achieve the desired task objectives across the different environments.


\textbf{Evaluation Results - Learning Performance:} We analyzed convergence times by referring to the learning curves in Figure \ref{fig:all-rewards}. Given that our approach and MAPLE use different affordances, then direct comparisons with MAPLE may not be appropriate since the reward functions are different. However, we can still assess convergence times, defined here as the time taken to learn a near-optimal policy for a given task.

In the Door environment, both our approach and MAPLE show approximately equal convergence times. For the Lift and Cleanup tasks, MAPLE converges slightly faster, possibly due to fewer primitive parameters and less exploration constraints from affordance coupling. In the Wipe task, our approach converges much faster, likely due to its ability to leverage variable stiffness, adapting force behavior to task requirements.

\begin{figure}[b!]
    \centering
    \includegraphics[width=0.9\columnwidth]{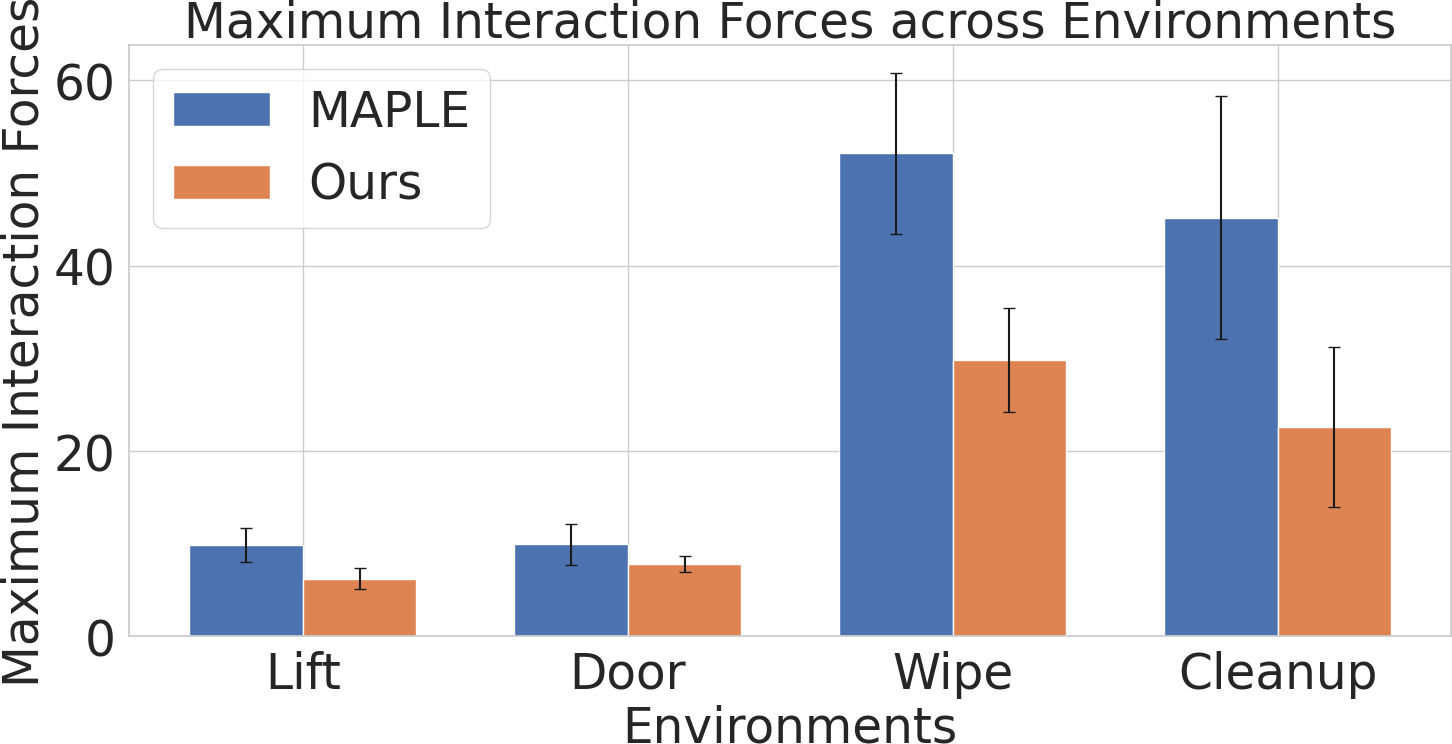}
    \caption{Comparison of maximum interaction forces}
    \label{fig:wrapfig}
    \vspace{-5mm} 
\end{figure}

\begin{figure*}[h!]
    \centering
    \includegraphics[width=\textwidth]{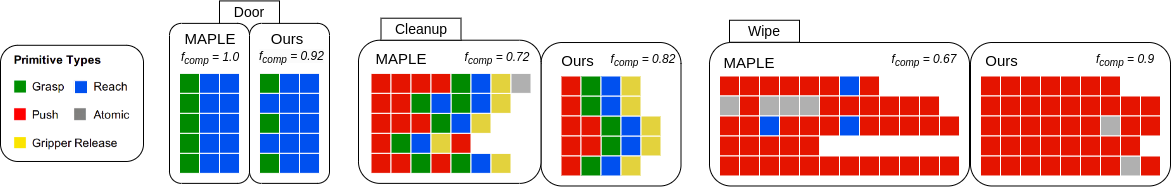}
    \caption{Compositionality comparison showcasing the learned sequential behavior. The rows correspond to primtive sequences generated by 5 sample environment runs.}
    \label{fig:compositionality}
    \vspace{-5mm}
\end{figure*}
\textbf{Evaluation Results - Maximum Interaction Force:} We demonstrate samples of the variable stiffness behavior across the different environments in Figure \ref{fig:stiff_profile}. We also include a graph showing the average applied end-effector forces over a sample of 500 evaluation runs in Figure \ref{fig:wrapfig}  highlighting our framework's ability to finish the task while exerting less force. These forces were acquired directly from the simulation environment.

In the Lift and Cleanup environments, both of which are tabletop settings, $K_z$ is maintained low when interacting near the table, while $K_x$ and $K_y$ are higher to ensure precise alignment with the objects of interest. In the Door environment, $K_x$ is relatively high to provide stability during initial contact, with $K_y$ increasing as the door handle is pushed down and all stiffness values decreasing when pulling the door open. In the Wipe environment, $K_x$ and $K_y$ are low since the primary action involves contact along the z-axis, while $K_z$ maintains a higher value to exert enough force for effective wiping without excessive interaction forces.





This increased compliance results in lower interaction forces across environments, as shown in Figure \ref{fig:wrapfig}. Our approach consistently exerts less force, with lower standard deviation, implying less sensitivity to task randomization. The Wipe and Cleanup tasks demonstrate this effect by showing a pronounced decrease in tabletop impact forces. Specifically, these forces are reduced when the robot slides an object along the surface during the Cleanup task and when it wipes away debris during the Wipe task. Note that the average force was only calculated across the successful trials in order to avoid biasing the results, since a robot not performing any actions generates no interaction forces.


\textbf{Evaluation Results - Compositionality:} We quantify recurring patterns of primitive choices for solving a given task using a \emph{compositionality metric} \cite{nasiriany2022augmenting}. A high compositionality score reflects the policy's ability to generate repeatable behavior sequences to complete a given task.

The compositionality was calculated for a sample of 30 successful environment runs, illustrated in Figure \ref{fig:compositionality}. The Lift task was excluded as it had the same compositionality score ($f_\text{comp}=1$), with a grasp and reach primitive sequence. In the Door task, we share the same number of primitive executions as MAPLE, but it shows more consistent primitive selection. In the Cleanup task, our approach reduces the number of primitive executions needed, likely due to more robust pushing and precise object approach. In the Wipe environment, our method has more consistent primitive selection than MAPLE, indicating better understanding of task requirements.

\textbf{Evaluation Results - Success Rate:} A comparison of success rates between MAPLE and our method is shown in Table \ref{table:combined_success}. Following training, the simulation and real world experiments were run 20 times to obtain the success rates. In the real world experiments, the policy was directly deployed onto the hardware with no fine-tuning to test the sim2real capabilities of the framework. MAPLE was not tested in real-world experiments due to its rigidity and potential operational hazards. Specifically, if the target state was defined at a location on or below the table surface, the robot's motion would lead to unintended force application and potentially cause damage to the environment.

\begin{table}[b!]
\centering
\caption{Success Rates (\%) for Simulation and Real World}
\label{table:combined_success}
\small 
\begin{tabular}{l@{\hspace{5pt}}cccc}
\hline
 & Lift & Door & Wipe & Cleanup \\
\hline
\specialcell{MAPLE \\(Simulation)} & \specialcell{100.0 \\$\pm$ 0.0 }& \specialcell{100.0 \\$\pm$ 0.0} & \specialcell{42.0 \\$\pm$ 11.7 }&\specialcell{ 91.0 \\$\pm$ 5.8}  \\ \hline
\specialcell{Ours \\(Simulation)} & \specialcell{100.0\\ $\pm$ 0.0} & \specialcell{100.0 \\$\pm$ 0.0} & \specialcell{86.0 \\$\pm$ 6.2} &\specialcell{ 87.0\\ $\pm$ 6.1 }  \\ \hline
\specialcell{Ours \\(Real World)} & 90.0 & - & 70.0 & 80.0 \\
\hline
\end{tabular}
\end{table}


Our approach achieves comparable success rates in Lift and Door tasks, while also improving the safety of the system due to its higher degree of compliance. In the Cleanup task, IMP-HRL achieves a slightly lower success rate than MAPLE. Given that our method prioritizes compliance, this highlights a tradeoff between safety and success in tasks requiring precise sequential object manipulation.

As for the Wipe task, our approach achieves double MAPLE's success rate. This significant improvement is attributed to our method's stiffness control capacity, as compared to MAPLE's use of position control. Position control works for Lift and Door but fails in wiping due to end-effector rigidity, leading to a misapplication of force or loss of contact. In contrast, our method ensures consistent surface contact and preventing excessive or insufficient force application.

\section{Conclusions and Limitations}
\label{sec:conclusion}
    This paper presents a hierarchical reinforcement learning framework aimed at enabling adaptive stiffness control in sequential contact tasks. It utilizes a pre-defined library of behavior primitives and equips them with variable stiffness capabilities. This was done by incorporating an expanded action space to allow the agent to modify its stiffness and an adaptive controller for dynamic stiffness modifications during primitive execution. During training, we introduce affordance coupling to combine position and stiffness affordances, which promotes efficient exploration while incentivizing compliance. The framework showcases notable results in learning efficiency, variable stiffness control, compositionality in primitive selection, and success rates when compared to MAPLE, a state-of-the-art framework in sequential planning. Furthermore, real-world evaluations validate the proposed approach's sim2real capability. 

The proposed method faces some limitations. The use of affordance coupling may limit learning efficiency when the task relies on accurate manipulation rather than contact or force interaction. This was evident in the experimental results for the Lift and Cleanup environments in which our method required more epochs to learn accurate manipulation. This can be attributed to the fact that affordance coupling incentivizes compliance, while manipulation tasks typically require some degree of stiffness to align the end-effector with a graspable object accurately. 
Another limitation lies in the acquisition of the adaptive stiffness controller parameters. Specifically, the controller relies on pre-defined scaling factors ($\beta$ and $\gamma$) that need to be set. They are acquired either through kinesthetic demonstrations, which require physical interaction with the robot, or iteratively tuning $\beta$ and $\gamma$ to match the desired performance, which can be time-consuming.



\bibliographystyle{IEEEtran}

\bibliography{references}  
\clearpage

\appendices

\section{Ablation Studies}\label{sec:appendix-ablation}
\begin{figure}[h!]
  \centering
    \centering
    \includegraphics[width=0.7\linewidth]{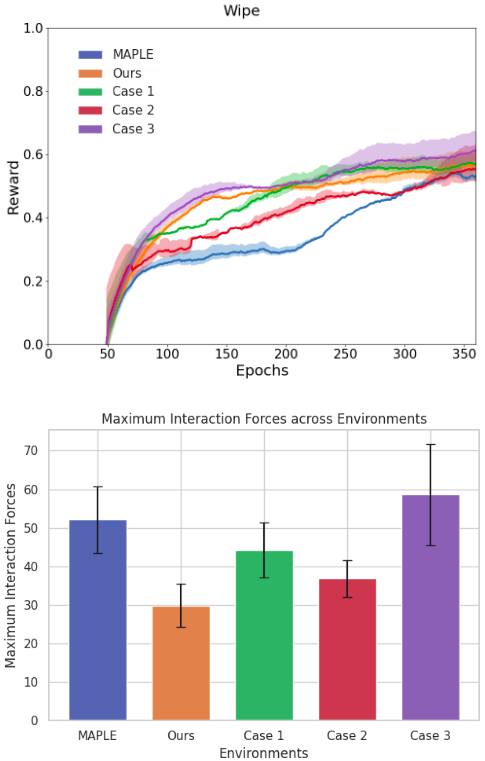}
  \caption{Comparison of convergence time and maximum interaction forces across 3 ablation cases}
  \label{fig:ablation}
\end{figure}

We conduct ablation studies to measure the impact of the added components on the performance of our system. Specifically, we trained a model on the Wipe environment due to the extensive contact nature of the task. Accordingly, we investigate 3 cases, each of which omits components of the proposed framework. The results are visualized in Figure \ref{fig:ablation} and are representative of the overall performance across all environments.
\begin{itemize}
    \item \underline{Case 1}: Extended action space \textit{with Adaptive Controller}
    \item \underline{Case 2}: Extended action space \textit{with Affordance Coupling}
    \item \underline{Case 3}: Extended action space \\
\end{itemize}
\textbf{Evaluation Results - Convergence Time.} The convergence time results in Figure \ref{fig:ablation} clearly reflect that the extension of the action space with stiffness parameters is the greatest contributor to the accelerated learning. On the other hand, the exclusive use of an adaptive controller (Case 1) or affordance coupling (Case 2) leads to a notable deterioration in learning performance, as compared to the use of both (Ours).


\textbf{Evaluation Results - Maximum Interaction Forces.} Figure \ref{fig:ablation} presents the maximum interaction forces achieved through variable stiffness in different frameworks. Our proposed approach consistently minimizes these forces during environmental interactions. This is followed by the framework using only an adaptive controller (Case 1), where stiffness reduction takes place in a relatively narrower range. In turn, this leads to a relatively higher force exertion. As for Case 3, it demonstrates that the exclusive reliance on an extended action space yields the worst performance due to a lack of incentive to reduce stiffness, so the agent opts for high stiffnesses to ensure stain removal.

Looking into Figure \ref{fig:ablation}, we can further infer that the use of affordance coupling (Case 2) leads to lower overall stiffnesses as compared to the adaptive controller (Case 1), demonstrated by the lower interaction forces. This further implies that incorporating the adaptive controller directly with MAPLE will not lead to lower maximum interaction forces, which is typically characterized by a lower overall stiffness.

\textbf{Evaluation Results - Stiffness Behavior.} Upon removing affordance coupling from the framework (Case 1), the agent exhibits a dependence on high stiffness values, which are subsequently reduced using the adaptive controller. As for the case that employs affordance coupling but omits the adaptive controller (Case 2), the agent tends to select relatively low stiffness values, but the profile remains static. Additionally, the absence of corrective behavior leads the agent to attempt corrections during the execution of the next primitive rather than concurrently with the current one. Additionally, removing both affordance coupling and the adaptive controller (Case 3) yields a static stiffness profile, and the agent tends to select high stiffness values due to a lack of incentive for reduction.


\section{Training \& Simulation}\label{sec:appendix-training}

\subsection{Training Setup}

The training codebase used is based on \textit{RLkit}\footnote{\url{https://github.com/rail-berkeley/rlkit}}, which in turn is based on \textit{rllab}\footnote{\url{https://github.com/rll/rllab}}. We document all the hyperparameters used in the training procedure in Table \ref{table:network-params} and \ref{table:training-params}. An important thing to add is that a target entropy is set for the first 200 epochs primarily to promote exploration for both the primitives and the stiffness parameters.

\begin{figure}[h!]
  \centering
  \includegraphics[width=\linewidth]{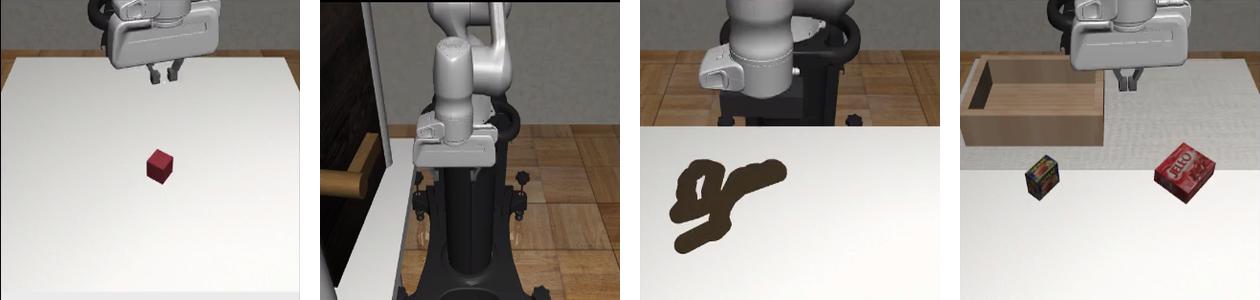}
    \caption{Simulation Environments: Lift, Door, Wipe, Cleanup}
  \label{fig:sim-pics-appendix}
\end{figure}

\begin{table*}
\centering
\caption{Network and Optimization Parameters}
\label{table:network-params}
\begin{tabular}{>{\centering\arraybackslash}p{6cm}|>{\centering\arraybackslash}p{4cm}}
\hline
\textbf{Hyperparameter} & \textbf{Value} \\
\hline
Network Structure (All Networks) & 512, 512 \\
Q network and policy activation & ReLU \\
Q network output activation & None \\
Policy network output activation & tanh \\
Optimizer & Adam \\
Batch Size & 1024 \\
Learning rate (all networks) & $3 \times 10^{-5}$ \\
Target network update rate $\tau$ & $1 \times 10^{-3}$ \\
\hline
\end{tabular}
\end{table*}

\begin{table*}
\centering
\caption{Training, Exploration, and Reward Factors}
\label{table:training-params}
\begin{tabular}{>{\centering\arraybackslash}p{6cm}|>{\centering\arraybackslash}p{4cm}}
\hline
\textbf{Hyperparameter} & \textbf{Value} \\
\hline
Discount Factor & 0.99 \\
Replay Buffer Size & $1 \times 10^6$ \\
Reward Scale & 5.0 \\
Affordance Score Scale $\lambda$ & 10.0 \\
Number of Training Steps per Epoch & 1000 \\
Number of Exploration Actions per Epoch & 3000 \\
Horizon Length per Episode & 150 actions (except wipe, 300) \\
\hline
\end{tabular}
\end{table*}

With regards to the observations used to train the model, the same observation is shared across all environments except Wipe. In those environments, the observations consist of:

\begin{itemize}
    \item Cartesian Pose
    \item Object Poses
    \item Distance from End-Effector to Object(s)
    \item Gripper State (either 0 or 1) \\
    \end{itemize}

\noindent As for the Wipe environment, the observation becomes:

\begin{itemize}
    \item Cartesian Pose
    \item Percentage Wiped
    \item Stain Centroid and Radius
    \item Distance from End-Effector to Centroid
\end{itemize}

\subsection{Simulation Setup}

Here, we specify the description of each task setup and specify their success conditions: \\





\underline{Lift}: \\
\textbf{Description}: There is a single cube on a tabletop \\
\textbf{Success Condition}: The cube is lifted above a height threshold (20 cm) \\

\underline{Door}: \\
\textbf{Description}: There is a hinged door with an L-handle \\
\textbf{Success Condition}: The handle exceeds a certain position (15 cm) and angle (30\textdegree) \\

\underline{Cleanup}: \\
\textbf{Description}: There is a jello box, a spam can, and a wooden box on a tabletop \\
\textbf{Success Condition}: The jello box is at a threshold distance from the table corner (10 cm) and the spam can is in a wooden box \\

\underline{Wipe}: \\
\textbf{Description}: There are stains on a tabletop, which are defined by their table coverage percentage (40\%) and stain line width (4 cm) \\
\textbf{Success Condition}: There are no stains on the table







\section{ Real World Experiments }\label{sec:appendix-realworld}

\begin{figure}[b!]
  \centering
  \includegraphics[width=0.7\linewidth]{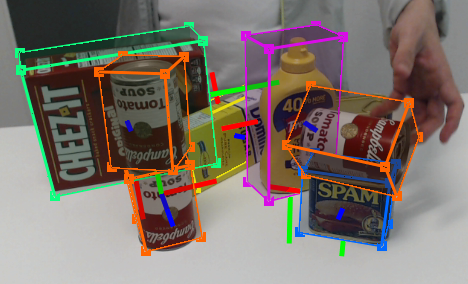}
  \caption{6D pose estimation of YCB object set \cite{tremblay2018deep}}
  \label{fig:sim-pics-appendix}
\end{figure}

This section discusses the experimental setup and evaluation of the Real World experiments.

\textbf{Hardware Setup.} An Intel RealSense D435i\footnote{\url{https://www.intelrealsense.com/depth-camera-d435i/}} was used to generate an RGB-D stream of the environment. These streams are later used to identify object poses. For the Lift and Cleanup experiments, we placed the camera on a tripod such that it is aligned with the tabletop. As for the Wipe environment, the camera was mounted on the robot end-effector to provide it with an accurate view of the stains.

\textbf{Software Setup.} ROS Noetic was used to interface between the cameras, trained model, and the robot. A RealSense ROS Wrapper\footnote{\url{https://github.com/IntelRealSense/realsense-ros}} was used to extract the RGB-D stream from the camera. In turn, we used Deep Object Pose \cite{tremblay2018deep} to estimate the 6D pose of the objects in the environment. Further details regarding observation acquisition are provided in Appendix~\ref{sec:appendix-realworld}.

\textbf{Robot Control.} The impedance controller used was the \textit{human-friendly controller}\footnote{\url{https://github.com/franzesegiovanni/franka_human_friendly_controllers}} \cite{franzese2021ilosa}. Since our model only outputs stiffness parameters and target positions, we used these parameters as input to the impedance controller. In turn, the controller acts as an interface with the robot to actuate its joints and reach the target position.

\textbf{Success Rate.} Each experiment was run 20 times with randomized object placements and end-effector starting positions. The model achieved a success rate of 90\% on the Lift task, 80\% on the Cleanup task, and 70\% on the Wipe~task.

As mentioned in Appendix \ref{sec:appendix-training}, the model uses object poses as part of the observation. In order to track the 6D pose of the environment objects in the real world, we use Deep Object Pose\footnote{\url{https://github.com/NVlabs/Deep_Object_Pose}} with the corresponding YCB objects\footnote{\url{https://ycbbenchmarks.com/object-set/}} used in simulation. It is important to note that the wiping task naturally does not involve interactions with solid objects, so we used a simple k-means clustering algorithm to identify the wiping stains based on color.

In the real-world experiments, the observations for the trained model were obtained using the Franka ROS Interface and the Intel RealSense D435i camera. More specifically, the process entails the following:

\begin{itemize}
    \item \textbf{Cartesian Pose} was extracted directly from the Franka ROS Interface. This information includes the position and orientation of the end-effector in the robot's workspace.
    \item \textbf{Gripper State} was extracted directly from the Franka ROS Interface. The gripper width was used to identify whether it was open or closed.
    \item \textbf{Object Pose} was estimated using Deep Object Pose with the Intel RealSense D435i camera. Using the RGB-D stream, Deep Object Pose analyzes the data and returns 6D object poses at a rate of 15 frames per second.
    \item \textbf{Distance from End-Effector to Object} was calculated directly given that we have both poses
\end{itemize}

As for the Wipe environment, there are two unique observation elements. First, K-Means clustering was used on the RGB stream, which was followed by color thresholding. This allows us to separate the black stains from the white background. Accordingly, the observations were acquired using the following methods:

\begin{itemize}
    \item \textbf{Percentage Wiped} was calculated using the initial stain as a template. By counting the number of black pixels, we can identify how many have been removed, which corresponds to the percentage wiped.
    \item \textbf{Stain Centroid and Radius} were acquired by pairing the RGB and Depth stream, which allows us to identify the location of the centroid and the radius of the stain.
\end{itemize}


\section{ Acquiring Adaptive Controller Parameters}\label{sec:appendix-acquiring-controller-parameters}

The \textbf{Adaptive Controller} adapts the stiffness in accordance with Equation  \ref{eq:Kdot}, in which $\beta$ and $\gamma$ are tunable parameters that influence the stiffness behavior. In this work, we acquire these parameters by collecting a sample of 15 kinesthetic demonstrations in which we equally perform reach, push, and atomic primitives. The gripper primitive was not demonstrated since it does not involve a change in stiffness. As for the grasp primitive, it consists of a reaching and grasping sequence, so the demonstration would be redundant.

The primitives consist of simple linear motions from the current state to a target state. Accordingly, the kinesthetic demonstrations consisted of moving the end-effector  move along a linear path from a random starting position towards a target state(s). We can then extract the stiffness profile used in these motions by referencing a method by Dou et al. \cite{kim2009impedance}, in which the impedance of a human arm is modeled as
\begin{equation}
    F_h = K_h \Delta x
\end{equation}
where $F_h$ is the interaction force and $K_h$ is the human arm stiffness. This force is then mapped to \textit{normalized stiffness} using
\begin{equation}
    K = \underline{K} + (\overline{K} - \underline{K}) \cdot \frac{(F_h - \underline{F}_h)}{(\overline{F}_h - \underline{F}_h)}
\end{equation}
where $\overline{K}$ and $\underline{K}$ represent the upper and lower thresholds of the calculated stiffness while $\overline{F}_h$ and $\underline{F}_h$ represent the upper and lower thresholds of the interaction force.

For each demonstration, we find the values of $\beta$ and $\gamma$ by minimizing the Mean Squared Error (MSE) between the demonstration $\dot{K}(t)$ values and the values predicted by the Equation \eqref{eq:Kdot}. Once applied across all the demonstrations, we average the acquired $\gamma$ and $\beta$ values and use them as parameters of the adaptive stiffness controller. All in all, this of parameter acquisition was performed only once, and allowed the controller to mimic the adaptive behavior presented by the demonstrator across all the evaluated tasks. Additionally, it saved a significant amount of time when compared to manually fine-tuning $\gamma$ and $\beta$ to acquire the desired behavior.

\end{document}